\documentclass[journal]{IEEEtran}

\usepackage{amsmath}
\usepackage{amssymb}
\usepackage{graphicx}
\usepackage{subcaption}
\usepackage[bottom]{footmisc}
\usepackage[ruled,vlined]{algorithm2e}
\usepackage{multirow}
\usepackage{xcolor}
\usepackage{enumitem}
\ifCLASSINFOpdf
\else
\fi

\begin{document}

\title{Cross-mode Knowledge Adaptation for \\ Bike Sharing Demand Prediction using \\ Domain-Adversarial Graph Neural Networks}

\author{Yuebing~Liang, Guan~Huang and Zhan~Zhao*
\thanks{Y. Liang and G. Huang are with Department of Urban Planning and Design, the University of Hong Kong, Hong Kong (e-mail: yuebingliang@connect.hku.hk; guanhuang@connect.hku.hk).}
\thanks{Z. Zhao is with Department of Urban Planning and Design, the University of Hong Kong, Hong Kong, and also with Musketeers Foundation Institute of Data Science, the University of Hong Kong, Hong Kong (zhanzhao@hku.hk) }
\thanks{* Corresponding author.}
}


\maketitle

\begin{abstract}
For bike sharing systems, demand prediction is crucial to ensure the timely re-balancing of available bikes according to predicted demand. Existing methods for bike sharing demand prediction are mostly based on its own historical demand variation, essentially regarding it as a closed system and neglecting the interaction between different transportation modes. This is particularly important for bike sharing because it is often used to complement travel through other modes (e.g., public transit). Despite some recent progress, no existing method is capable of leveraging spatiotemporal information from multiple modes and explicitly considers the distribution discrepancy between them, which can easily lead to negative transfer. To address these challenges, this study proposes a domain-adversarial multi-relational graph neural network (DA-MRGNN) for bike sharing demand prediction with multimodal historical data as input. A temporal adversarial adaptation network is introduced to extract shareable features from demand patterns of different modes. To capture correlations between spatial units across modes, we adapt a multi-relational graph neural network (MRGNN) considering both cross-mode similarity and difference. In addition, an explainable GNN technique is developed to understand how our proposed model makes predictions. Extensive experiments are conducted using real-world bike sharing, subway and ride-hailing data from New York City. The results demonstrate the superior performance of our proposed approach compared to existing methods and the effectiveness of different model components.  
\end{abstract}

\begin{IEEEkeywords}
bike sharing; demand prediction; inter-modal relationships; graph neural networks; adversarial learning
\end{IEEEkeywords}

\IEEEpeerreviewmaketitle

\section{Introduction}

\IEEEPARstart{B}ike sharing is an emerging sustainable, convenient and generally affordable mode of transportation. Over the past decade, it has been widely deployed in many cities around the world as an effective solution to traffic congestion and last-mile problems. Due to its positive impact on the environment and public health, bike sharing systems (BSSs) are being promoted to play an increasingly important role in urban transportation systems. However, the efficient operation of BSSs is challenged by fluctuations in spatial and temporal demand patterns, resulting in inefficient bike repositioning and high operating costs for bike rebalancing \cite{xu2018station}. Accurate short-term demand prediction at high spatial resolution is the basis to support dynamic rebalancing of available bikes and ensure the user experience of bike sharing service.

Numerous studies have been conducted for the problem of bike sharing demand prediction. Early attempts use regression or machine learning models to solve this problem, which suffer from relatively low accuracy. Recently, more research interests have shifted to deep learning methods, due to their demonstrated effectiveness in extracting the complex knowledge hidden in large-scale mobility data \cite{xu2018station, jin2020dockless}. In particular, Graph Neural Networks (GNNs) have been employed in the demand prediction problem and achieved state-of-the-art performance \cite{geng2019spatiotemporal,yu2017spatio}. Despite the success of these methods, they regard bike sharing as a closed system and neglect the potential rich information of the interaction between BSS and other transportation modes. This is especially important to consider for bike sharing, because it is mainly used for short-distance trips or the first-mile/last-mile portion of longer trips. In practice, BSS are often designed as feeders to public transit systems or support multimodal transportation connections \cite{zhang2018short}. As a result, the demand for bike sharing will inevitably be influenced by other transportation modes, which should be considered in demand prediction. Incorporating demand information across modes can also help mitigate the data sparsity problem commonly seen in BSS, since bike sharing is rarely one of the primary travel modes in cities.

To incorporate inter-modal relationships, several recent studies have investigated the joint demand prediction of multimodal transportation systems using multi-task learning frameworks \cite{ye2019co,wang2020multi,xu2022adaptive}. These methods treat different modes equally and might not be effective enough when we take interest in a particular mode. Another group of research focused on the demand prediction of a target mode by adapting the learned knowledge from other modes \cite{zhang2018short, cho2021enhancing,li2021multi}. This study belongs to the second category, in which we aim to enhance the prediction performance of BSS with the help of cross-mode demand information. Despite existing relevant works, there still exist several challenges to be resolved.

The first challenge is how to fuse spatiotemporal information from stations or zones of multiple modes. To utilize multimodal demand information, existing methods either aggregate the usage of other modes as additional attributes of BSS stations for model input \cite{zhang2018short, cho2021enhancing, lv2021mobility}, or capture mode-specific temporal dependencies with recurrent-based models and transfer the learned temporal features across modes with knowledge adaptation techniques \cite{li2021multi,hua2022transfer}. They do not consider the spatial dependencies between spatial units of different modes and are not capable of leveraging spatiotemporal knowledge hidden in the multimodal system. Although recent research has introduced graph neural networks to capture cross-mode spatial relationships \cite{xu2022adaptive,liang2021joint}, they are designed for jointly predicting the demand of multiple modes and do not directly consider knowledge adaptation issues, such as distribution discrepancy. 

The second challenge is how to handle the distribution discrepancy between the demand patterns of different modes. For instance, our preliminary analysis shows that there is a notable disparity between the temporal patterns of bike sharing, subway and ride-hailing in Manhattan: generally, bike sharing usage is more active in the daytime, while the demand of subway is more concentrated during rush hours and ride-hailing is more busy at night (see Section~\ref{experiment:data} for more details). This is not uncommon in multimodal transportation systems, as people use different modes for various purposes. Directly leveraging multimodal data without reducing the effect of distribution discrepancy can easily lead to negative transfer. This has been confirmed in previous research \cite{cho2021enhancing}, in which they find simply incorporating subway demand data as model input has negative effect on the bike demand prediction result.

The third challenge is how to understand the mechanism regarding why the model makes such prediction. Although deep neural networks (DNNs) have demonstrated great capabilities in prediction performance, they have long been criticized for lack of interpretability. Recently, explainable AI techniques have been employed in transportation research for explaining human mobility behavior, such as destination or route choices \cite{simini2021deep, zhao2022deep}. However, the spatiotemporal relationships in multimodal transportation systems are much more complicated and remain to be uncovered from "black-box" models.

To address these issues, in this research, we propose a domain-adversarial multi-relational graph neural network (DA-MRGNN) for bike sharing demand prediction based on cross-mode knowledge adaptation. 
We define a multi-relational graph containing multiple intra- and inter-modal graphs to encode the spatial dependencies between stations or zones from different modes, and use a temporal adversarial adaptation network and a multi-relational graph neural network \cite{liang2021joint} to extract cross-mode temporal and spatial relationships respectively. 
An explainable GNN technique is introduced to uncover the learnt characteristics of the multimodal transportation network. Experiments are conducted using the Citi Bike data from New York City (NYC), with the subway and ride-hailing data as additional inter-modal demand inputs. Because of data constraint, we will focus on station-based BSS in this study, but the methodology should be generalizable to stationless (or dockless) BSS, as well as other similar mobility systems. The specific contributions of this research are as follows:

\begin{itemize}[noitemsep]
    \item We propose a domain-adversarial graph neural network for bike sharing demand prediction based on cross-mode relationships learnt from historical demand data. With an adapted multi-relational graph neural network, the proposed approach can effectively fuse spatiotemporal information from stations or zones of different modes.
    \item To reduce distribution discrepancy between modes, we introduce a temporal adversarial adaptation network which learns transferable features from auxiliary modes to BSS stations with adversarial learning.
    \item We develop an explainable multi-relational GNN technique, which extends GNNExplainer \cite{ying2019gnnexplainer} to multi-relational graph structures. It is capable of identifying important stations or zones that influence the prediction of a BSS station in the multimodal network.
    \item Extensive experiments are conducted based on real-world datasets from NYC, and the results demonstrate the effectiveness of our proposed model and all the introduced components.
\end{itemize}

This study extends a preliminary version of the proposed model \cite{liang2022bike} in the following aspects: (1) with the adversarial loss to promote positive knowledge adaptation across modes, the performance of our proposed model is improved (see Section~\ref{method: TAAN}); (2) with the newly introduced explainable GNN technique, we provide more domain insights regarding the spatial dependencies between stations and zones in the multimodal network (see Section~\ref{method:interpret} and Section~\ref{res:MR-GNNExplainer}); (3) this work provides more comprehensive experimental results to evaluate how the effects of cross-mode information vary over space and time (in Section~\ref{res:modes}), and how the proposed components impact the prediction performance (in Section~\ref{res:ablation}).

\section{Related Work}\label{Related Work}

In this section, we first review existing methods regarding bike sharing demand prediction, and then provide a brief summary of domain adaptation and explainable GNNs, which are relevant to our methodology.

\subsection{Bike sharing demand prediction}

Traditional methods for bike sharing demand prediction mainly focus on finding the relationships between bike demand and its historical demand 
through regression models, such as auto-regressive integrated moving average (ARIMA) \cite{yoon2012cityride} and linear regression (LR) \cite{rudloff2014modeling}. Later research uses other more advanced machine learning models. Feng et al. \cite{feng2018hierarchical} introduced a hierarchical demand prediction model based on gradient boosting regression tree. Guidon et al. \cite{guidon2020expanding} estimated the demand of bike sharing services using spatial regression models and random forest. These methods may not be adequate to capture the complex relationships between input variables and bike demand, and thus are not precise enough.

In recent years, extensive studies have presented the power of deep learning models to capture the nonlinear and complex relationships for demand prediction tasks. Xu et al. \cite{xu2018station} adopted a long-short term memory neural network (LSTM) with exogenous factors as additional input. Jin et al. \cite{jin2020dockless} applied temporal convolution networks (TCNs) to capture temporal dependencies for bike sharing demand prediction. 
To incorporate spatial information, previous research employed convolutional neural networks (CNNs) with recurrent networks \cite{zhou2018predicting, li2021multi}, which however, require aggregating demand data to a grid system. 
Graph neural networks (GNNs) exempt the requirement of artificial segmentation, which is more useful for BSS operations. Lin et al. \cite{lin2018predicting} proposed a graph convolutional network (GCN)-based approach along with recurrent networks for station-level bike demand prediction. A purely convolutional model was introduced in \cite{yu2017spatio} with GCNs and TCNs to model spatial and temporal dependencies respectively. 
To capture more complex correlations hidden in the transportation network, a multi-graph convolutional neural network was proposed in \cite{geng2019spatiotemporal}, using multiple graphs to encode different types of spatial dependencies. Wu et al. \cite{wu2019graph} proposed to learn spatial dependencies directly from traffic data, with an adaptive adjacency matrix learned through node embedding. However, these methods are mode-specific and do not consider inter-modal relationships.

To leverage the cross-mode demand information, increasing attention has been paid to the co-prediction of multiple transport modes. In \cite{ye2019co} and \cite{wang2022nonlinear}, the demands for taxis and bike sharing are aggregated to a grid system to enable shareable feature learning, before co-predicted using a convolutional recurrent network. 
For demand prediction of general multimodal systems with diverse spatial units, Xu et al. \cite{xu2022adaptive} introduced a co-modal Graph Attention Network to capture the interactions between different modes. A multi-relational graph neural network (MRGNN) was developed in our previous work \cite{liang2021joint} for the joint demand prediction of subway and ride-hailing. These methods pre-assume that different transport modes have similar demand patterns and might not be effective when the distribution of different modes are significantly different. In addition, they treat different transport modes equally and may not work well when we take specific interest in a target mode. 

Though limited, several recent works have exploited cross-mode demand information for bike sharing demand prediction. Zhang et al. \cite{zhang2018short} developed an LSTM-based model with the historical demand of bus and subway as additional input. A graph learning approach was introduced in \cite{cho2021enhancing} to enhance the prediction accuracy of BSS demand during peak hours with public transit usage information. Lv et al. \cite{lv2021mobility} focused on the demand prediction of BSS stations around subway stations by extracting subway-related features. These methods process the historical flow of adjacent public transit stations as additional attributes of BSS stations, and do not model the relationships between stations or zones from different modes directly. To capture temporal dependencies across modes, a memory-augmented recurrent model was proposed in \cite{li2021multi} for the demand prediction of a station-sparse mode by taking advantage of the demand information of a station-intensive mode. Hua et al. \cite{hua2022transfer} enhanced the prediction performance of bike sharing with a pre-trained LSTM model for public transit using transfer learning strategies. These methods only consider temporal correlations across modes, while a model that can effectively transfer knowledge across modes considering the effect of spatial dependencies between heterogeneous spatial units is needed. 

\subsection{Domain adaptation}

Domain adaptation is a sub-field within transfer learning that aims to reduce the effect of distribution discrepancy when transferring knowledge across domains \cite{farahani2021brief}. To learn transferable representations, previous works focus on minimizing the distribution distance between domains \cite{liu2021knowledge}. Tzeng et al. \cite{tzeng2014deep} introduced a domain confusion loss based on Maximum Mean Discrepancy (MMD). A Deep Adaptation Network was developed in \cite{long2015learning} which reduces the multi-kernel MMD of embedded representations from different domains in a reproducing kernel Hilbert space. Recently, inspired by the idea of adversarial learning, domain adversarial neural networks (DANN) was introduced in \cite{ganin2015unsupervised}, which learns domain-invariant feature representations that cannot be distinguished by a domain discriminator. Later research demonstrate the effectiveness of domain adversarial learning for multiple transfer learning tasks in the field of computer vision and natural language processing, such as image classification, face recognition and language understanding \cite{farahani2021brief}.

Several recent studies have introduced the idea of adversarial learning to cross-domain human mobility prediction. Tang et al. \cite{tang2022domain} proposed an adversarial spatiotemporal network for short-term traffic forecasting across cities. A spatial adversarial adaptation module was designed in \cite{fangtransfer} to capture transferable spatial correlations between source and target cities. Wang et al. \cite{wang2020multi} developed a multi-task learning framework for crowd-level and OD-level flow prediction, with an adversarial loss for shared feature extraction across tasks. These methods aimed at transferring knowledge across cities or tasks, which are different from this paper, where we use domain adversarial networks to mitigate the distribution discrepancy between multiple transport modes.  

\subsection{Explainable Graph Neural Networks}

While GNNs have been a powerful tool for human mobility prediction, the complex graph structure has made it difficult to understand its working mechanism. Meanwhile, the ability to understand GNN's prediction is crucial for improving model's transparency and providing domain insights regarding graph characteristics. As a result, increasing efforts have been devoted for explainable GNNs. GNNExplainer \cite{ying2019gnnexplainer} is one of the pioneering works, whose main idea is to identify a subgraph pattern that is important for the prediction of the model. XGNN \cite{yuan2020xgnn} formulated a reinforcement learning-based graph generation task which aims to generate graph structures that can mostly recover GNN's prediction. PGExplainer \cite{luo2020parameterized} utilized a generative probabilistic model to provide a global understanding of predictions made by GNNs. For a comprehensive review of existing explainable GNN techniques, we refer readers to \cite{li2022survey}. However, the aforementioned explainable GNN techniques are all developed for a single homogeneous graph, and are not directly applicable to the multi-relational graph defined in our study. In addition, most of them are experimented with synthetic, molecule or social networks, while how to understand human mobility networks with explainable GNN techniques is underexplored.

\section{Problem Statement}\label{Problem}
In this section, we introduce some notations in this research and then formulate our problem.  

\textit{Definition 1 (Demand Sequence):} Consider a transport mode $m$ with $N_m$ nodes (i.e. stations/service zones). For each node $i = 1, 2, ..., N_m$, its inflow and outflow demand at time step $t$ is denoted as $x_{m, i}^t\in \mathbb{R}^2$. Next, we represent the demand of all the nodes from mode $m$ at time step $t$ as $X_m^t=\{x_{m,0}^t, x_{m,1}^t…, x_{m,N_m}^t\},X_m^t \in \mathbb{R}^{N_m \times 2}$. Further, we use $X_m^{t-T:t}=\{X_m^{t-T}…, X_m^{t-1},X_m^{t}\}$ to denote the demand sequence of mode $m$ over time steps $T$.

\textit{Problem (Bike Sharing Demand Prediction):} This research aims to predict the station-level bike sharing demand given historical demand of BSS as well as other modes. Formally, given the historical demand of bike sharing, denoted as $X_{b}^{t-T:t}$, and that of auxiliary modes, i.e., subway and ride-hailing in our case, denoted as $X_{s}^{t-T:t}$ and $X_{h}^{t-T:t}$, the goal is to predict bike sharing demand $X_{b}^{t+1}$ at the next time step : 
\begin{equation}
X_{b}^{t+1}=F(X_{b}^{t-T:t},X_{s}^{t-T:t},X_{h}^{t-T:t}),
\end{equation}
where $F(*)$ is the prediction function to be learned by our proposed model. This formulation can be easily adapted to other demand prediction problems with multimodal historical demand as input.

\section{Methodology}\label{method}
\begin{figure*}[ht!]
  \centering
  \includegraphics[width=\linewidth]{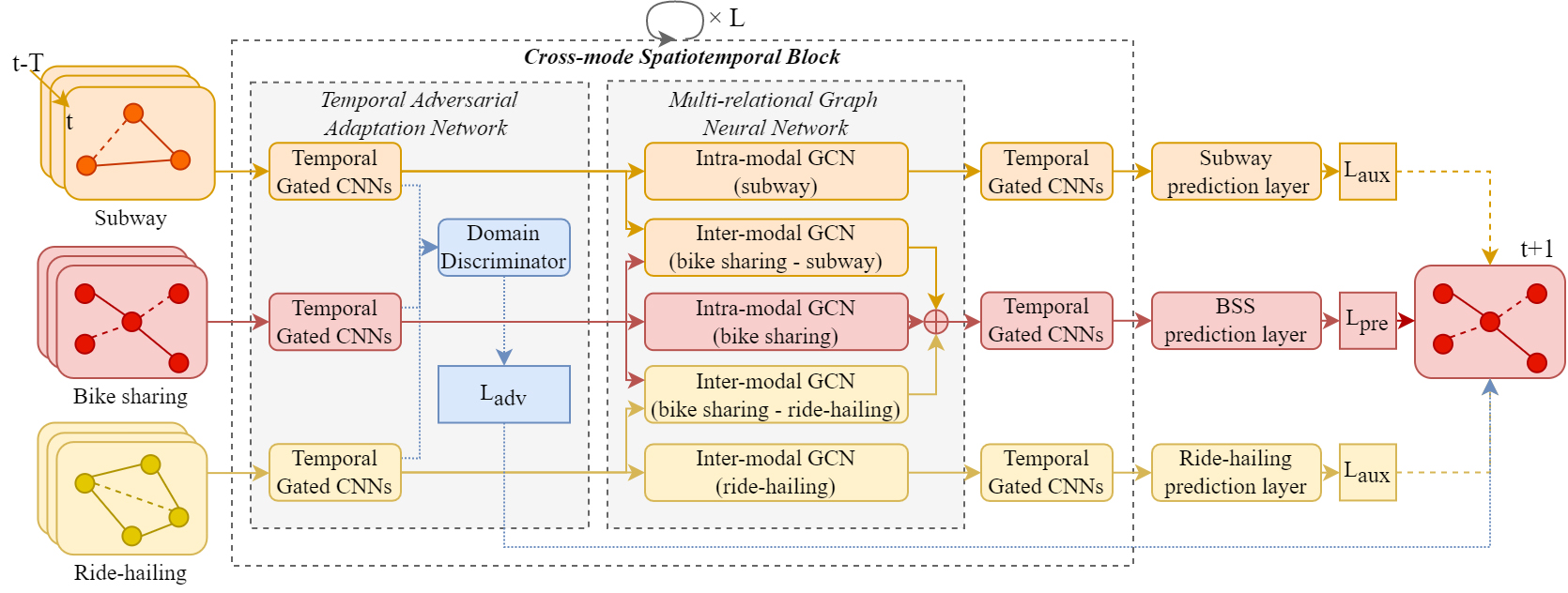}
  \caption{The architecture of DA-MRGNN 
  }
  \label{fig:ST-MRGNN}
\end{figure*}

This section presents a domain-adversarial multi-relational graph neural network (DA-MRGNN) for station-level bike sharing demand prediction by taking advantage of historical demand sequences of subway and ride-hailing systems. Fig.~\ref{fig:ST-MRGNN} displays the overall architecture of our proposed model. It is composed of several stacked cross-mode spatiotemporal blocks (ST-blocks) for multimodal knowledge extraction, each comprising a temporal adversarial adaptation network (TAAN) to capture shareable temporal features across modes and a multi-relational graph neural network (MRGNN) \cite{liang2021joint} to model the spatial influence of adjacent subway stations and ride-hailing zones on BSS stations. Based on the learned representation from ST-blocks, a prediction layer is used to generate bike sharing demand prediction. In the prediction layer, we use the demand prediction of subway and ride-hailing as auxiliary tasks for model training, which can enhance the prediction performance. We further introduce an explainable GNN technique, namely multi-relational GNNExplainer (MR-GNNExplainer), to understand how our model makes predictions. Details of each module are introduced below.

\subsection{Temporal adversarial adaptation network}\label{method: TAAN}
To capture temporal correlations within the historical flow of multiple modes, a naive approach would be to apply a separate temporal modeling component for each mode. The learned temporal features will then be fused in a MRGNN layer which will be introduced later. However, our preliminary analysis shows that there is a notable disparity between the temporal patterns of different modes and neglecting the effect of cross-mode pattern discrepancy might lead to negative transfer (see Section~\ref{experiment:data} for more details).
To address this issue, we design a temporal adversarial adaptation network (TAAN) by taking advantage of adversarial learning \cite{ganin2015unsupervised}. 
Specifically, it is composed of a temporal feature extractor for each mode to capture temporal dependencies and a domain discriminator to distinguish the learned features from different modes. The feature extractors are trained to confuse the domain discriminator so that they learn mode-transferable feature representations.

\subsubsection{Temporal feature extractor}

Previous research typically used RNNs or CNNs to extract temporal features for mobility prediction. In this research, we employ a gated convolution network (TCN) proposed by \cite{yu2017spatio} as the temporal feature extractor for each mode, due to its fast training time and simple structures. Briefly, given an input sequence, it uses a 1-D causal convolution to capture the relationships between each time step and its neighborhoods, along with an output gate to control the ratio of information that passes through layers. In our case, a separate TCN layer is applied to each mode to capture mode-specific temporal information, which processes all nodes in the mode in parallel. Mathematically, for a node $i$ in mode $m$, the input of the TCN layer is a sequence of feature vectors denoted as $h_{m, i} \in \mathbb{R}^{K \times c}$, where $K$ is the length of the sequence and $c$ is the dimension of the input vector. In the first ST-block, the input feature sequence is the historical demand series, i.e., $h_{m, i} = x_{m, i}^{t-T:t}, h_{m, i} \in \mathbb{R}^{T \times 2}$. Given $h_{m, i}$, the TCN layer is formulated as:
\begin{equation}
h_{m, i}^{(c)} = (W_{c,1}^{(m)} \star h_{m, i} + b_{c,1}^{(m)}) \odot \sigma(W_{c,2}^{(m)} \star h_{m, i} + b_{c,2}^{(m)}),
\end{equation}
where $\star$ denotes the convolution operation, $\odot$ is the element-wise multiplication and $\sigma(\ast)$ represents the sigmoid function. The model parameters $W_{c,1}^{(m)}$, $b_{c,1}^{(m)}$ are used for feature transformation of nodes in mode $m$ and $W_{c,2}^{(m)}$, $b_{c,2}^{(m)}$ are used to compute the output gate for mode $m$. $h_{m, i}^{(c)} \in \mathbb{R}^{K' \times c'}$ is the learned representations for node $i$ in mode $m$ from the TCN layer, where $K'$ is the length of the output feature sequence and $c'$ is the dimension of the output vector.  

\subsubsection{Domain discriminator}

The domain discriminator is formulated as a multi-class classifier, which uses the learned representation from the TCN layer as input and maps it to a probability vector denoting which mode the representation is from. Specifically, we first mix the learned representations from different modes. For each node $i$ of mode $m$, the learned feature matrix from the TCN layer $h_{m, i}^{(c)}$ is then flattened to a 1-d vector, denoted as $\overline{h}_{m, i}^{(c)}$. Given $\overline{h}_{m, i}^{(c)}$, we predict the mode it comes from using a feed-forward network:
\begin{gather}
d_{i,1} = ReLU(W_{d,1} \overline{h}_{m, i}^{(c)} + b_{d,1}), \\
d_{i,2} = ReLU(W_{d,2} d_{i,1} + b_{d,2}), \\
\hat{d_i} = softmax(W_{d,3} d_{i,2} + b_{d,3}), 
\end{gather}
where $W_{d,1}, W_{d,2}, W_{d,3}$ are the parameter matrices for linear transformation and $b_{d,1}, b_{d,2}, b_{d,3}$ are the biased terms. The output $\hat{d_i} \in \mathbb{R}^{3}$ is a predicted probability vector denoting which mode the node $i$ is from. The domain discriminator is optimized by minimizing the cross-entropy loss:
\begin{equation}
L_{adv} = -\sum_{m \in \{b, s, h\}}{\sum_{i \in m}o_{m,i} log(\hat{d_i})},
\end{equation}
where $o_{m,i}$ is a one-hot vector denoting the real mode that the node $i$ is from. Meanwhile, the temporal feature extractors should be trained to maximize $L_{adv}$ in order to learn transferable temporal features across modes. This is achieved by inserting a gradient reversal layer (GRL) \cite{ganin2015unsupervised} between the TCN layers and the domain discriminator in each ST-block. During backward propagation, GRL reverses the gradient from the domain discriminator and passes it to the preceding TCN layers. In this way, the temporal feature extractors and the domain discriminator can be optimized simultaneously with simple backpropagation.

\subsection{Multi-relational graph neural network}\label{method:MRGNN}
\begin{figure}[ht!]
  \centering
  \includegraphics[width=\linewidth]{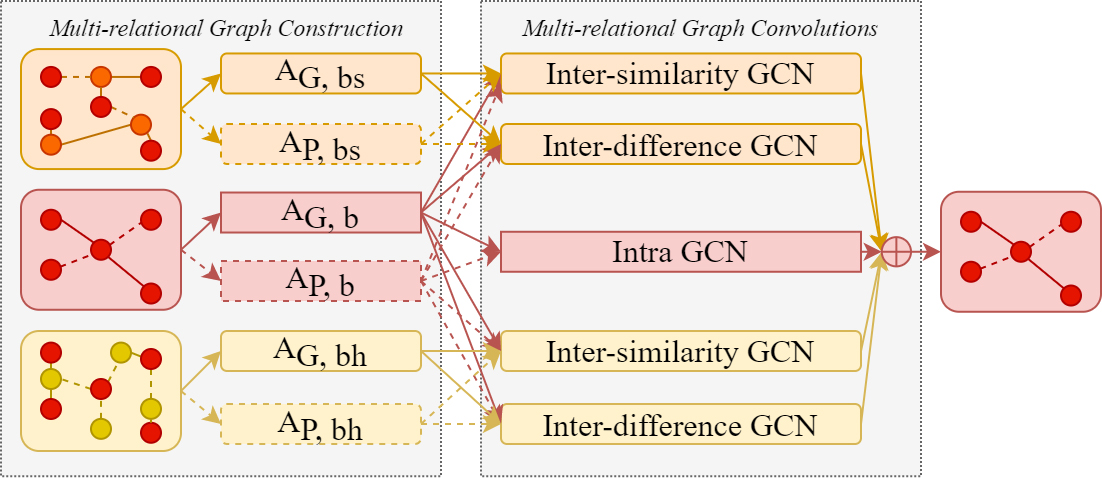}
  \caption{The framework of MRGNN}\label{fig:MRGNN}
\end{figure}

In this subsection, we provide more details on MRGNN that is used to capture the interactions between spatial units across modes. As shown in Fig.~\ref{fig:MRGNN}, MRGNN is composed of two major parts: \textit{multi-relational graph construction} to encode cross-mode spatial dependencies and \textit{multi-relational graph convolutions} to capture correlations between nodes through message passing.

\subsubsection{Multi-relational graph construction}\label{method: graph_construct}
As subway and BSS (in our case) are station-based and ride-hailing is stationless, it is difficult to model spatial dependencies on a single homogeneous graph. To encode spatial dependencies within and across modes, we define multiple intra- and inter-modal graphs. Specifically, \textbf{intra-modal graphs} are used to capture spatial correlations among stations/zones of the same mode. Taking bike sharing as an example, its intra-modal graph is defined as $G_{b} = (V_b, A_b)$, where $V_b$ is a set of BSS stations, and $A_b \in \mathbb{R}^{N_b \times N_b}$ is an adjacency matrix representing the spatial dependencies between BSS stations. Similarly, an intra-modal graph is defined for subway and ride-hailing, denoted as $G_{s}$ and $G_{h}$.
\textbf{Inter-modal graphs} are defined to capture the pairwise correlations among stations/zones between the target mode (i.e., bike sharing) and each of the auxiliary modes (i.e., subway and ride-hailing). For example, the inter-modal graph between bike sharing and subway is represented as $G_{bs}=(V_b, V_s, A_{bs})$, where $V_b$ and $V_s$ denote BSS and subway stations and $A_{bs} \in \mathbb{R}^{N_b \times N_s}$ is a weighted matrix indicating the cross-mode dependencies between adjacent BSS and subway stations. Similarly, an inter-modal graph is defined between bike sharing and ride-hailing denoted as $G_{bh}$. 

Prior studies have demonstrated that strong correlations may occur between locations that are either geographically close or semantically similar (i.e., demand patterns) \cite{geng2019spatiotemporal}. To encode both relationships, we define two adjacency matrices for each graph: one for \textbf{geographical proximity} denoted as $A_G$ and, the other for \textbf{pattern similarity} denoted as $A_P$. The specific definitions of $A_G$ and $A_P$ follow our prior work \cite{liang2021joint}.

\begin{figure}[ht!]
  \centering
  \includegraphics[width=\linewidth]{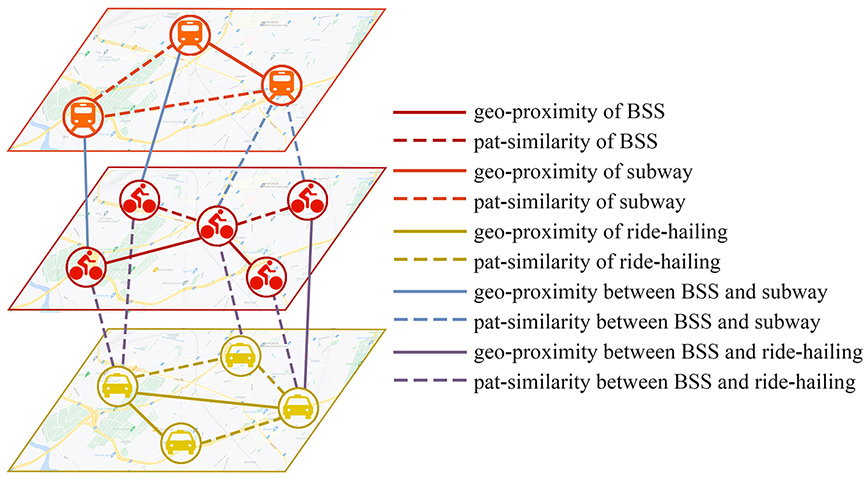}
  \caption{Modeling multimodal spatial dependencies for BSS}\label{fig:graph construct}
\end{figure}

Fig.~\ref{fig:graph construct} illustrates the constructed multi-relational graph of bike sharing, subway and ride-hailing. A total of (3+2)x2=10 relations is defined to encode spatial dependencies between nodes from different modes, including 3 intra-modal and 2 inter-modal graphs, each with 2 adjacency matrices. This formulation can be easily adapted to encode spatial dependencies across other modes with diverse network structures. 

\subsubsection{Multi-relational graph convolutions}\label{graph_convolution}

Graph convolutions have been an effective way to aggregate information of connected nodes. However, most existing GCNs cannot be applied to the multi-relational graph defined above for two main reasons: first, they cannot process inter-modal graphs with different types of nodes and a non-square adjacency matrix; second, they fail to consider the pattern difference between stations and zones from different modes. To tackle these issues, we introduce an inter-modal graph convolution network considering both cross-mode similarity and difference. It takes the learned representations of stations and zones from the TAAN module as input and aggregates information of connected subway stations and ride-hailing zones for each BSS station considering the effect of spatial dependencies. Mathematically, for each time step, given the inter-modal graphs $G_{bs}=(V_b, V_s, A_{bs})$ and $G_{bh}=(V_b, V_h, A_{bh})$, as well as the representations of subway and ride-hailing learned from the TCN layers, denoted as $H_s^{(c)} \in \mathbb{R}^{N_s \times c'}$, $H_h^{(c)} \in \mathbb{R}^{N_h \times c'}$ respectively, the \textbf{inter-modal similarity} is modeled as:
\begin{gather}
Z_{bs}^{(c)} =  ReLU({\widetilde{A}_{b}} (\widetilde{A}_{bs}H_s^{(c)}) W_{bs}^{(c)}+l_{bs}^{(c)}),\\
Z_{bh}^{(c)} =  ReLU({\widetilde{A}_{b}} (\widetilde{A}_{bh}H_h^{(c)}) W_{bh}^{(c)}+l_{bh}^{(c)}),
\end{gather}
where $Z_{bs}^{(c)} \in \mathbb{R}^{N_b \times c''}$, $Z_{bh}^{(c)} \in \mathbb{R}^{N_b \times c''}$ are the aggregated features of connected subway stations and ride-hailing zones on BSS stations respectively, $c''$ represents the output vector dimension of each node (i.e. station/zone) from the GCN layer. $W_{bs}^{(c)}, W_{bh}^{(c)}$ and $l_{bs}^{(c)}, l_{bh}^{(c)}$ are the learned model parameters. ${\widetilde{A}} = \frac{A}{rowsum(A)}$ denotes the normalized adjacency matrix constructed from $A$. 

Although we have extracted mode transferable temporal features in the TAAN module, they only mitigate the pattern discrepancy between modes and do not consider the heterogeneity of pairwise pattern distribution difference between stations and zones. Inspired by \cite{zhou2021modeling}, we use an \textbf{inter-modal difference} GCN to model such node-level distribution difference between bike sharing and the two auxiliary modes:
\begin{gather}
Z_{bs}^{(d)} =  ReLU({\widetilde{A}_{b}}|\widetilde{A}_{bs}H_s^{(c)} - H_b^{(c)}| W_{bs}^{(d)}+l_{bs}^{(d)}),\\
Z_{bh}^{(d)} =  ReLU({\widetilde{A}_{b}}|\widetilde{A}_{bh}H_h^{(c)} - H_b^{(c)}| W_{bh}^{(d)}+l_{bh}^{(d)}),
\end{gather}
where $|\widetilde{A}_{bs}H_s^{(c)} - H_b^{(c)}|$, $|\widetilde{A}_{bh}H_h^{(c)} - H_b^{(c)}|$ represent the information gap between BSS stations and the aggregated information of connected subway stations and ride-hailing zones respectively, $W_{bs}^{(d)}, W_{bh}^{(d)}, l_{bs}^{(d)}, l_{bh}^{(d)}$ are the learned model parameters.

In addition, we apply a standard GCN layer to model pairwise correlations between connected BSS stations. Given the intra-modal graph $G_b$ and the representations of bike learned from the TCN layer $H_b^{(c)} \in \mathbb{R}^{N_b \times c'}$, the correlations between BSS stations are modeled as:
\begin{equation}
{Z}_{b} =  ReLU({\widetilde{A}_{b}} H_b^{(c)} {W}_{b}^{(c)}+l_{b}^{(c)}),
\end{equation}
where ${W}_{b}^{(c)}, l_{b}^{(c)}$ are model parameters. In practice, the inter-modal similarity and difference graph convolution layers as well as the intra-modal graph convolution layer can be modeled in parallel using batch matrix multiplication operations. Through the intra- and inter-modal graph convolutions, each BSS station receives multiple feature vectors from geographically adjacent or semantically similar subway stations, ride-hailing zones and other BSS stations. The learned feature vectors from heterogeneous neighborhood nodes are then aggregated using an adding function.   

\subsection{Cross-mode spatiotemporal block} \label{method:st-block}
In this subsection, we describe the design of ST-blocks, which are stacked to capture correlations between nodes of different modes from both spatial and temporal domains. As introduced above, each ST-block uses a TAAN module to extract shareable temporal features from subway stations, ride-hailing zones and BSS stations. The learned features of different modes from TCN layers are then fed into the inter-modal graph convolution layer to jointly model spatial and temporal dependencies of BSS stations on connected subway stations and ride-hailing zones. We also apply an intra-modal graph convolution layer on each mode to capture spatiotemporal correlations within modes. Through experiments, we use an additional TCN layer for each mode after the MRGNN layer which can improve the model performance. This is potentially because the additional convolution layer promotes fast-state propagation between graph convolutions \cite{yu2017spatio}. The input of the second TCN layer is the sum of the output of TAAN and MRGNN, as a residual connection function to speed up model training. To stabilize model parameters, we employ a layer normalization function at the end of each ST-block. After several stacked ST-blocks, each BSS station gets a learned representation vector that summarizes the spatiotemporal information from multimodal historical demand. In addition, each subway station and ride-hailing zone is represented by a vector summarizing intra-modal demand information. 

\subsection{Prediction layer} \label{method:represent_sum}
Based on the learned representations from the stacked ST-blocks, we generate bike sharing demand predictions at the next time step using a  
fully-connected feed-forward network. The prediction error is given by:
\begin{equation}
\label{eq:pre_loss}
L_{pre} = \sum_{i \in b}||\hat{x}_{b,i}^{t+1}-{x}_{b,i}^{t+1}||^2,
\end{equation}
where $\hat{x}_{b,i}^{t+1}, {x}_{b,i}^{t+1}$ are the predicted and true demand values of BSS station $i$ at time step $t+1$. 

To train our model more efficiently, we incorporate the idea of auxiliary task learning \cite{liebel2018auxiliary} in the prediction layer. Briefly, different from the main task which generates the target output for an application, auxiliary tasks are not directly relevant to the application, but can enhance the performance of the main task by assisting with feature extraction of the input data. In our case, we use the future predictions of subway and ride-hailing demand as auxiliary tasks. The prediction is generated with additional feed-forward networks using the learned representations from the ST-blocks as input. The intuition behind it is that by jointly optimizing the demand prediction of auxiliary modes, we can guide the model to extract useful spatiotemporal information from them, which can in turn benefit bike sharing demand prediction. Essentially, the prediction error of subway and ride-hailing demand is used as an additional regularization term in the loss function: 
\begin{equation}
\label{eq:aux_loss}
L_{aux} = \sum_{i \in s}||\hat{x}_{s,i}^{t+1}-{x}_{s,i}^{t+1}||^2 + \sum_{j \in h}||\hat{x}_{h,j}^{t+1}-{x}_{h,j}^{t+1}||^2,
\end{equation}
where $\hat{x}_{s,i}^{t+1}, {x}_{s,i}^{t+1}$ are the predicted and real demand of subway station $i$, $\hat{x}_{s,i}^{t+1}, {x}_{s,i}^{t+1}$ are the predicted and real demand of ride-hailing zone $j$ at time step $t+1$. The final loss function contains three parts: prediction loss of BSS demand $L_{pre}$, adversarial loss of the domain discriminator $L_{adv}$ and prediction loss of the demand of auxiliary modes $L_{aux}$. The overall loss function is defined as follows:
\begin{equation}
\label{eq:loss}
L = L_{pre} + \epsilon_{aux} L_{aux} + \epsilon_{adv} L_{adv},
\end{equation}
where $\epsilon_{aux}, \epsilon_{adv}$ are used to trade-off different parts of loss. 


\subsection{Multi-relational GNNExplainer}\label{method:interpret}

In addition to making accurate predictions for bike sharing demand, people are also interested in why the model makes such prediction and whether it can provide domain insights regarding the relationship between BSS stations and other transportation modes. In this research, we interpret our proposed model by extending an explainable GNN technique named GNNExplainer \cite{ying2019gnnexplainer}. For a node in the graph, GNNExplainer identifies a subgraph that maximizes its mutual information with the GNN's prediction. 
The underlying assumption is that if the prediction result of the target node is largely determined by a connected node, their dependencies should be strong. The subgraph is identified by formulating a mean field variational approximation and learning a real-valued graph mask to select important edges. Mathematically, consider a general graph $G=\{V, A\}$, where $V$ is a set of nodes and $A \in \mathbb{R}^{|V| \times |V|}$ is an adjacency matrix. For the interpretation of a node $i \in V$, GNNExplainer first defines a computation graph $G_i=\{V_i, A_i\}$, where $V_i \subseteq V$ contains nodes in the neighborhood of $i$ and $A_i \in \mathbb{R}^{|V_i| \times |V_i|}$ denotes the connectivity of $V_i$. The subgraph $G_{s,i}$ that is important for the prediction of node $i$ is learned using the following optimization framework:
\begin{equation}
\min_{M_i}{dist(F_i(X_i, G_i), F_i(X_i, G_{s,i}=A_i \odot \sigma(M_i)))},
\end{equation}
where $M_i \in \mathbb{R}^{|V_i| \times |V_i|}$ is the mask to be learned, $X_i$ is the input features of nodes $V_i$, $\sigma(*)$ denotes the sigmoid function that maps $M_i$ to binary values, $\odot$ denotes element-wise multiplication, $F_i(*)$ denotes the model generating predicted outputs for node $i$ and $dist(*)$ is a distance function to measure the difference between the prediction result using the entire computation graph $G_i$ and the subgraph $G_{s,i}$.

While GNNExplainer has successfully explained simple graphs such as molecule and social networks, they cannot be directly applied to the multimodal transport systems in our research due to the complex multi-relational graph structure containing multiple intra- and inter-modal graphs. Although it is possible to add a mask on the adjacency matrices of each graph, our preliminary experiments show that such setting cannot achieve stable and satisfying results, likely due to the large number of parameters to be learned. 
To address this issue, we introduce multi-relational GNNExplainer (MR-GNNExplainer), which makes two modifications: firstly, instead of adding masks to the adjacency matrix, we add masks to nodes, which can greatly decrease model parameters; 
secondly, we construct a node mask for each mode in the multimodal network so that our proposed framework can identify important intra- and inter-modal relationships simultaneously. Mathematically, for a target BSS station $i$, we randomly initialize a node mask for bike sharing, subway and ride-hailing, respectively denoted as $M_{b,i} \in {\mathbb{R}^{N_b}}, M_{s,i} \in {\mathbb{R}^{N_s}}, M_{h,i} \in {\mathbb{R}^{N_h}}$. The masks are optimized using:
\begin{gather}
\begin{aligned}
\min_{\{M_{b,i}, M_{s,i}, M_{h,i}\}} dist(F_i(X_b^{t-T:t}, X_s^{t-T:t}, X_h^{t-T:t}), \\F_i(X_{b,i}^{t-T:t}, X_{s,i}^{t-T:t}, X_{h,i}^{t-T:t})),
\end{aligned} \\
X_{b,i}^{t-T:t} = X_b^{t-T:t} \odot \sigma(M_{b,i}), \\
X_{s,i}^{t-T:t} = X_s^{t-T:t}\odot \sigma(M_{s,i}), \\
X_{h,i}^{t-T:t} = X_h^{t-T:t}\odot \sigma(M_{h,i}).
\end{gather}
We use L2 norm as the distance function $dist(*)$. The learned node masks can effectively select a small subset of subway stations, ride-hailing zones and BSS stations that are important to the GNN's prediction of a target BSS station. For the interpretation results in our case, please refer to Section~\ref{res:MR-GNNExplainer}.

\section{Experimental Setup}\label{Experiments}

\subsection{Data Description}\label{experiment:data}
Our proposed model is validated on real-world public datasets from NYC. In this study, we use Manhattan as the research area and collect travel demand data of bike sharing, subway and ride-hailing from 2018-03-01 to 2018-08-31. Specifically, we use the following three datasets: 
\begin{itemize}
    \item \textbf{NYC Citi Bike}: the data consists of the pick-up and drop-off time and station of each Citi Bike trip records. During our study period, there are around 36 thousand trip records per day. We filter out the stations with an average of fewer than three orders per hour and keep 246 BSS stations for demand prediction.
    \item \textbf{NYC Subway}: the data contains the entries and exits counts of each turnstile in subway stations every four hours, with around 2.4 million entry/exit counts every day on average. We filter out subway stations with no demand for long time periods, which results in 107 subway stations.
    \item \textbf{NYC Ride-hailing}: we use the for-hire vehicle (FHV) data from NYC Taxi \& Limousine Commission (TLC), which is provided by ride-hailing companies such as Uber and Lyft. It contains the pick-up and drop-off time and zone of each trip records during the study period. On average there are 234 thousand trips per day during our study period. The zones are pre-determined by TLC and there are 63 TLC zones in Manhattan.
\end{itemize}

\begin{figure}
    \centering
    \includegraphics[width=\linewidth]{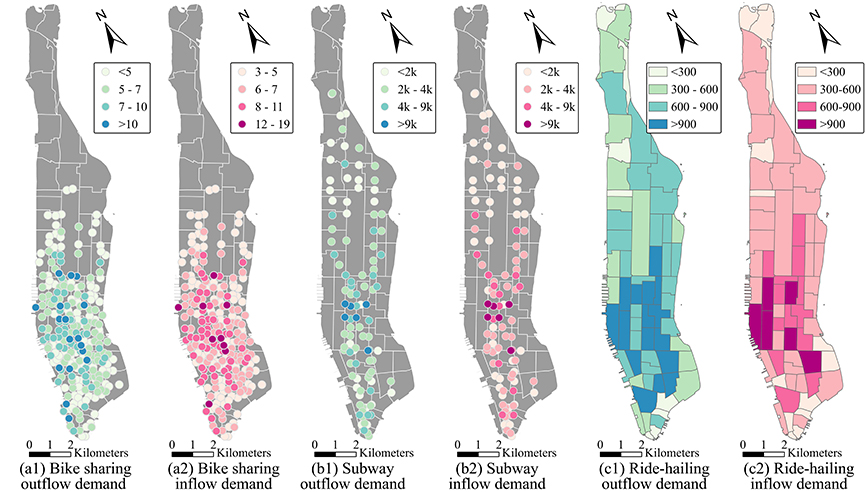}
    \caption{Spatial distribution of multimodal travel demand in Manhattan}
    \label{fig:od_s}
\end{figure}

\begin{figure}
    \centering
    \includegraphics[width=\linewidth]{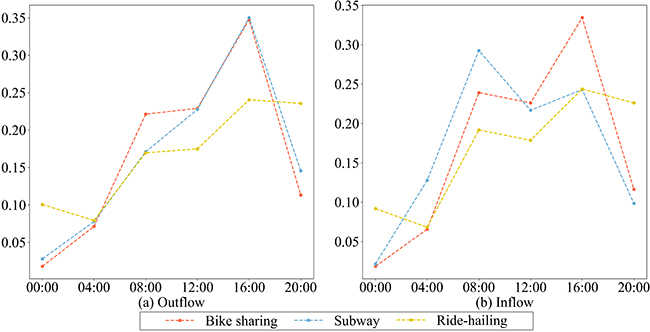}
    \caption{Temporal pattern of multimodal travel demand in Manhattan. (The travel demand of each mode is normalized for more intuitive comparison.)}
    \label{fig:od_t}
\end{figure}

We present the spatial distribution and temporal pattern of the multimodal travel demand in NYC in Fig.~\ref{fig:od_s} and Fig.~\ref{fig:od_t}. 
For all modes, stations/zones with intense demand are concentrated in Midtown Manhattan, followed by Downtown. This indicates that the demand of different modes exhibits similar spatial distributions, supporting our idea of using inter-modal demand to enhance the prediction performance of bike sharing. From Fig.~\ref{fig:od_t}, we find that there is a notable disparity between the temporal patterns of different modes. The usage of bike sharing is more active during 08:00-20:00 especially at evening peak (16:00-20:00), while ride-hailing is busier at night (16:00-24:00). On the other hand, the outflow and inflow demand of subway exhibits different patterns: its inflow has two peaks in the morning and evening, while its outflow is concentrated in the evening peak. This is likely because most people take the subway to Manhattan for work in the morning and leave Manhattan for home in the evening. Such result supports the necessity to mitigate the effect of different temporal patterns when leveraging cross-mode information. 

\subsection{Experiment settings}
We compare DA-MRGNN with two groups of baselines. The first group contains two methods with its own historical demand as input:
\begin{itemize}
    \item \textbf{XGBoost}: a machine learning model capturing the relationship between future demand and historical demand series based on gradient boosting trees;
    \item \textbf{LSTM}: a variant of RNNs using memory units to capture long-term temporal dependencies; 
    \item \textbf{STGCN} \cite{yu2017spatio}: a framework using GCNs for spatial dependencies and TCNs for temporal dependencies; 
    \item \textbf{MGCN} \cite{geng2019spatiotemporal}: a GCN-based model using multiple graphs to encode spatial correlations and conceptual RNNs to model temporal correlations; 
    \item \textbf{GWNET} \cite{wu2019graph}: a graph learning approach which captures the hidden spatial dependencies in the demand data by learning an adjacency matrix through node embedding. 
\end{itemize}

The second group contains BSS demand prediction methods considering the effect of other modes: 
\begin{itemize}
    \item \textbf{MM-STGCN} \cite{cho2021enhancing}: a STGCN-based approach which incorporates the usage of auxiliary modes as additional features of BSS stations for model input. For each BSS station, the usage features of subway and ride-hailing are computed as a weighted sum of the usage record of subway stations and ride-hailing zones, weighted by distance to the BSS station. 
    \item \textbf{MM-GWNET}: Following the idea of MM-STGCN, we implement a GWNET-based model with the accumulated demand of subway stations and ride-hailing zones as additional input.
\end{itemize}

For fair comparison, we use the same experiment settings for all models. Specifically, the demands of all three modes are aligned into 4-hour intervals and min-max normalization is used for each mode to mitigate the effect of demand variance. We set the historical time step $T=6$ and predict demand at the next time step. Data from the first 60\% time steps are used for model training, the following 20\% for validation, and the last 20\% for model testing. For all DNN-based models, we use 500 training epochs, with the learning rate of 0.002, the batch size of 32 and the dropout ratio of 0.3. To prevent overfitting, we use early stopping on the validation set and a L2 regularization on the loss function with a weight decay of 1e-5. All models are evaluated using three widely used metrics: Root Mean Square Error (RMSE), Mean Absolute Error (MAE) and Coefficient of Determination ($R^2$). We run 10 independent experiments for each model and report the average values on the test set. 

\section{Results}
\subsection{Comparison of model performance} \label{res:performance}

\begin{figure}[ht!]
  \centering
  \includegraphics[width=\linewidth]{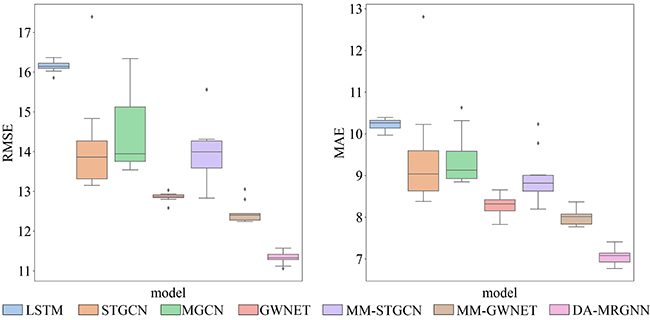}
  \caption{Comparison of model stability}\label{fig:boxplot}
\end{figure}

In this subsection, we compare the prediction performance of our proposed model with baselines on NYC bike sharing data. The performance of different models are summarized in Table~\ref{table:perform_compare}. Compared with the baseline models, our proposed model achieves significantly superior performance for all evaluation metrics. This is likely because our model can directly leverage spatiotemporal information of related subway stations and ride-hailing zones to enhance the prediction of bike sharing demand. In addition, our model reduces the effect of distribution discrepancy between modes with adversarial learning.  
Fig.~\ref{fig:boxplot} presents the performance variance of 10 independent runs of our proposed model and DNN-based baselines. We can find that our proposed model has the best performance regarding both RMSE and MAE in all experiments with relatively high stability. 

\begin{table}[h]
\caption{Performance comparison with baseline models}
\label{table:perform_compare}
\begin{center}
\begin{tabular}{c c c c c}
\hline
Input & Models & RMSE & MAE & $R^2$\\
\hline
\multirow{5}{*}{Single-mode} & XGBoost & 15.139 & 9.262 & 0.734\\
& LSTM & 16.148 & 10.255 & 0.695\\
& STGCN & 13.820 & 9.097 & 0.789\\
& MGCN & 14.436 & 9.359 & 0.760\\
& GWNET & 12.689 & 8.092 & 0.814\\
\hline
\multirow{3}{*}{Multi-mode} & MM-STGCN & 14.078 & 8.953 & 0.780 \\
& MM-GWNET & 12.461 & 7.998 & 0.820\\
& \textbf{DA-MRGNN} & \textbf{11.334} & \textbf{7.054} & \textbf{0.851}\\
\hline
\end{tabular}
\end{center}
\end{table}

Among baseline models, 
The poor performance of LSTM can be potentially explained by its inability to capture spatial dependencies between BSS stations. XGBoost performs notably better than LSTM, showing the advantage of gradient boosting machines in some cases.  Although MGCN encodes multiple types of spatial dependencies, it performs worse than STGCN in our experiments, which might be due to its low model stability as illustrated in Fig.~\ref{fig:boxplot}. GWNET performs better than the other GCN-based baselines given its ability to learn spatial dependencies hidden in demand data with an adaptive adjacency matrix. Compared with STGCM, MM-STGCN performs even worse regarding RMSE and $R^2$ in our case. MM-GWNET performs better than GWNET, though the improvement is marginal. This suggests that incorporating multi-modal knowledge as additional attributes of BSS stations might not be able to make significant contributions to the model performance and may even lead to negative transfer. Compared with MM-GWNET, our proposed model can further reduce the prediction error, with RMSE and MAE improvement of 9.0\% and 11.8\% respectively.

\subsection{The effect of input modes}\label{res:modes}

To further investigate the effect of incorporating inter-modal relationships on bike sharing demand prediction, we compare the performance of our model using different input mode combinations: bike sharing alone, bike sharing and subway, bike sharing and ride-hailing, and all three modes. The results are shown in Table~\ref{table:mode_combine}. Note that they share the same model structure, just with different input modes. We find that the inter-modal relations from geographically nearby or semantically similar subway stations and ride-hailing zones can indeed help with the prediction of bike sharing demand: without any inter-modal relations, the prediction error regarding RMSE and MSE would increase by 10.3\% and 11.9\% respectively. Incorporating either subway or ride-hailing demand patterns can already significantly improve the prediction performance of bike sharing demand, with the RMSE reduced by 6.7\% and 7.0\% respectively. It is also worth noting that with only bike sharing demand as input, our proposed model can still perform better than the single-mode version of GWNET, suggesting that our spatiotemporal framework can better extract intra-modal relationships. 

\begin{figure}
    \includegraphics[width=\linewidth]{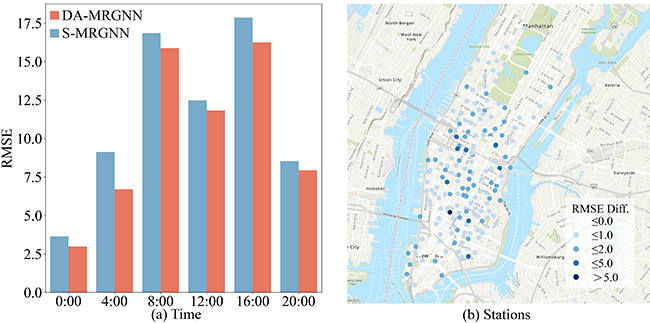}
    \caption{Comparison of DA-MRGNN and S-MRGNN (a variant of DA-MRGNN with single-mode input) at different time and stations}
    \label{fig:rmse_diff}
\end{figure}

To investigate how the effect of inter-modal relationships varies over space and time, we compare our proposed model with a variant of DA-MRGNN using single-mode demand as input (S-MRGNN) at different time and stations. The results are displayed in Fig.~\ref{fig:rmse_diff}. It can be clearly seen from Fig.~\ref{fig:rmse_diff} that DA-MRGNN performs better than S-MRGNN at all time intervals. The advantage of DA-MRGNN over S-MRGNN is especially significant during 04:00-08:00, indicating that our model can effectively utilize the knowledge from other modes to improve the prediction accuracy of demand-sparse time periods. During 16:00-20:00, DA-MRGNN also performs notably better than S-MRGNN. This suggests that our model can successfully capture the complex relations between different modes during rush hours. Fig.~\ref{fig:rmse_diff}b shows that the prediction performance of almost all stations can benefit from the cross-mode knowledge, especially for several BSS stations in Midtown and Downtown Manhattan. In summary, inter-modal relations can contribute to the demand prediction performance of bike sharing across different time and space.

\begin{table}[h]
\caption{Performance comparison of different mode combinations}
\label{table:mode_combine}
\begin{center}
\begin{tabular}{c c c c}
\hline
Models & RMSE & MAE & $R^2$\\
\hline
Bike sharing & 12.498 & 7.890 & 0.822\\
Bike sharing + subway & 11.656 & 7.326 & 0.845\\
Bike sharing + ride-hailing & 11.619 & 7.314 & 0.847\\
Bike sharing + subway + ride-hailing & \textbf{11.334} & \textbf{7.054} & \textbf{0.851}\\
\hline
\end{tabular}
\end{center}
\end{table}

\subsection{The effect of model components}\label{res:ablation}
In this subsection, we conduct ablation experiments to quantify the contribution of different components in our proposed model. We drop different components of DA-MRGNN to construct model variants and compare their performance with the full model. The components are listed below:

\begin{itemize}
    \item Adversarial loss ($L_{adv}$): In the TAAN layer, we utilize adversarial learning to guide the model to learn transferable features from auxiliary modes. With adversarial learning ablated, the domain discriminator module is dropped and the model is trained with only $L_{pre}$ and $L_{aux}$. 
    \item Inter-modal difference GCNs (diff-GCNs): our proposed inter-modal GCN uses two GCN layers to model cross-mode similarity and difference respectively. In this variant, the inter-modal GCN only captures the similarity between stations and zones from different modes.
    \item Auxiliary loss ($L_{aux}$): In the prediction layer, we incorporate the demand prediction of subway and ride-hailing as auxiliary tasks. Without auxiliary tasks, the model is optimized with only $L_{pre}$ and $L_{adv}$.
\end{itemize}

\begin{figure}
    \centering
    \includegraphics[width=\linewidth]{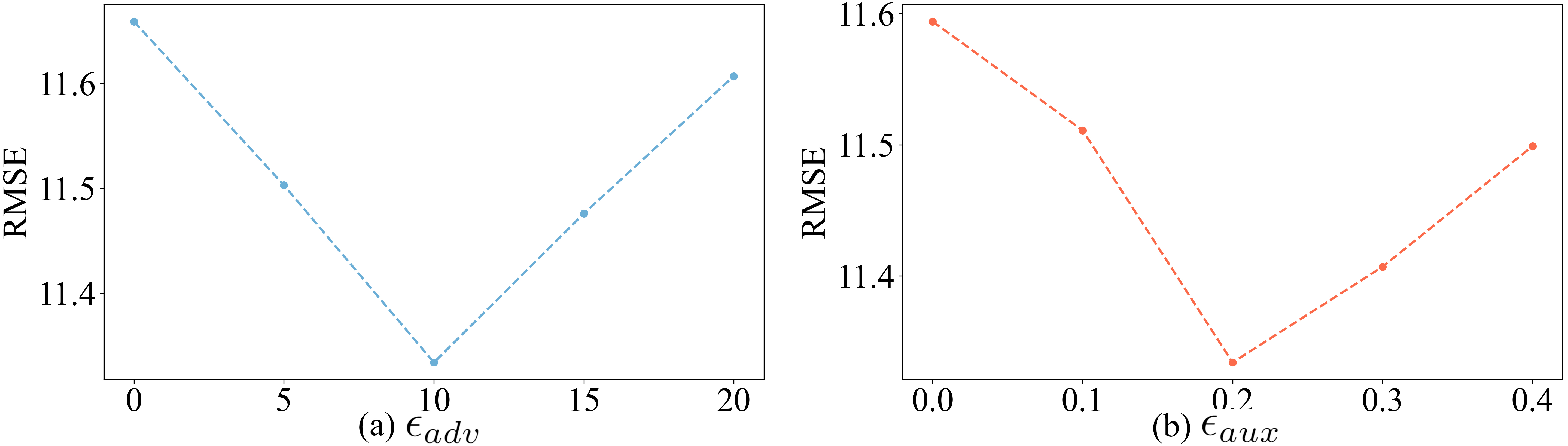}
    \caption{The influence of $\epsilon_{adv}$ and $\epsilon_{aux}$ on model performance}
    \label{fig:trade_off}
\end{figure}

Table~\ref{table:ablation} shows the averaged performance of different model variants based on 10 independent runs. We can see that removing any one of them will lead to notable increase in the prediction error. We further use a T-test to evaluate the performance improvement provided by different components and it is found the differences between model variants and the full model are all statistically significant with $p$-value small than 0.05. This confirms that the listed components can all improve our model performance significantly. To further investigate how the trade-off between $L_{pre}$, $L_{adv}$ and $L_{aux}$ influences the model performance, we change $\epsilon_{adv}$ from 0 to 20 with a step size of 5 and $\epsilon_{aux}$ from 0.0 to 0.4 with a step size of 0.1. The results are displayed in Fig.~\ref{fig:trade_off}. It is found that the increase of $\epsilon_{adv}$ can help reduce the prediction error when $\epsilon_{adv} < 10$, whereas the performance degrades as $\epsilon_{adv}$ gets larger. 
This indicates that, rather than completely eliminating the distribution discrepancy between modes, it can be helpful to preserve some meaningful inter-modal differences. Also, a large $\epsilon_{adv}$ may lead to the model over emphasizing on $L_{adv}$, which can be detrimental to the overall prediction performance. When $\epsilon_{aux}$ is smaller than 0.2, the model performs relatively poor likely because it is unable to extract useful representation from auxiliary modes, which then in turn degrades the prediction performance of BSS demand. In addition, the model performance is less satisfying with a high $\epsilon_{aux}$ which is reasonable as the model takes less account into the optimization of BSS demand prediction. 

\begin{table}[h]
\caption{Performance comparison of model variants (averaged over 10 runs)}
\label{table:ablation}
\begin{center}
\begin{tabular}{c c c c}
\hline
Models & RMSE & MAE & $R^2$\\
\hline
- $L_{adv}$ & 11.659 & 7.273 & 0.845\\
- diff-GCNs & 11.527 & 7.243 & 0.847 \\
- $L_{aux}$ & 11.594 & 7.217 & 0.847\\
DA-MRGNN & \textbf{11.334} & \textbf{7.054} & \textbf{0.851}\\
\hline
\end{tabular}
\end{center}
\end{table}

\subsection{Interpretation analysis using MR-GNNExplainer}\label{res:MR-GNNExplainer}

\begin{figure}
    \centering
    \includegraphics[width=\linewidth]{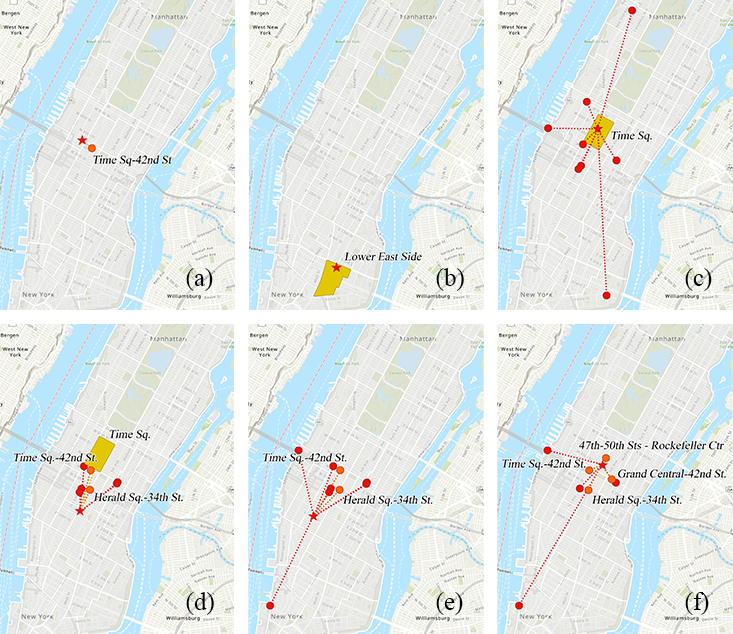}
    \caption{Exemplar explanation subgraphs for bike sharing demand prediction. In each subfigure, the red star denotes the predicted BSS station. The red points, orange points and yellow zones denote connected BSS stations, subway stations and ride-hailing zones that are important to its prediction.}
    \label{fig:local_interpret}
\end{figure}

To further understand how DA-MRGNN utilizes intra- and inter-modal relationships to make predictions, we develop an explainable GNN technique namely MR-GNNExplainer as introduced in Section~\ref{method:interpret}. Specifically, it provides local explanation for the prediction of each BSS station by identifying a subset of connected subway stations, ride-hailing zones and BSS stations that are important to its prediction. Fig.~\ref{fig:local_interpret} shows the interpretation results for a few BSS stations. It can be seen that the BSS station in Fig.~\ref{fig:local_interpret}a is mainly affected by an adjacent subway station. Similarly, the prediction result of the BSS station in Fig.~\ref{fig:local_interpret}b is mainly determined by the ride-hailing zone it falls in. This verifies the effectiveness of considering spatially adjacent subway stations and ride-hailing zones for bike sharing demand prediction. In addition, BSS stations can also be highly affected by distant subway stations or ride-hailing zones. For example, the BSS station in Fig.~\ref{fig:local_interpret}d is highly affected by the ride-hailing zone of Time Square, which however is not in its neighborhood. This is potential because distant stations/zones can also have similar pattern distributions or strong OD connections, which are useful for the demand prediction. 

\begin{figure}
    \centering
    \includegraphics[width=\linewidth]{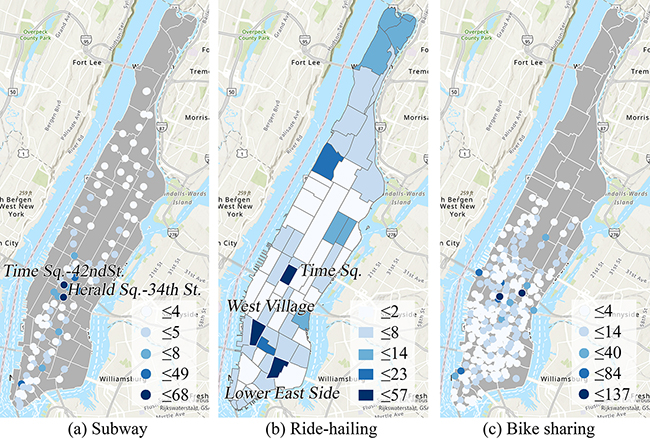}
    \caption{The importance score of stations and zones in the multimodal system.}
    \label{fig:global_interpret}
\end{figure}

To provide a global understanding of how stations/zones from different transport modes influence the multimodal system, we define an importance score for each station and zone. Specifically, we first generate a subgraph (i.e., a subset of nodes in our case) for each BSS station using MR-GNNExplainer, and then count the appearance frequency of each subway station, ride-hailing zone and BSS station in all the subgraphs. The underlying assumption is that if a station or zone has high influence on more BSS stations, it might play a more important role in the multimodal system. The results are displayed in Fig.~\ref{fig:global_interpret}. It can be found that within each mode, there are several stations or zones that are much more influential than the other ones. As shown in Fig.~\ref{fig:global_interpret}a, \textit{Time Sq-42nd St} and \textit{Herald Sq-34th St} are the two most important subway stations. As to ride-hailing zones, \textit{Time Sq}, \textit{West Village} and \textit{Lower East Side} play the most important role. Two bike sharing stations in Midtown Manhattan are the most influential among BSS stations. Through further investigation, we find that what these stations and zones have in common is their high outflow to many different destinations. These results suggest that a multimodal system might be dominated by several "key" nodes with intense and scattered demand.

\section{Conclusion}\label{Conclusion}
Bike sharing demand prediction is quite crucial for the efficient operation of bike sharing systems. In this paper, we aim to enhance the prediction performance of bike sharing by incorporating cross-mode information. This problem is challenging regarding how to fuse multimodal spatiotemporal information and how to handle the distribution discrepancy of temporal patterns between modes. To address these issues, a domain-adversarial graph neural network is proposed to extract shareable spatiotemporal features from spatial units of different modes. To extract shareable temporal features, a temporal adversarial adaptation network is developed by taking advantage of adversarial learning. The spatial dependencies across modes are encoded with multiple intra- and inter-modal graphs, and an inter-modal graph convolution layer is introduced to capture both correlations and difference between nodes from different modes. In addition, to better understand how our model makes prediction, we design an explainable GNN technique, which is capable of making local explanations by identifying an important subset of connected nodes. Empirical validation is performed on real-world bike sharing, subway and ride-hailing datasets from NYC. The results show that our proposed model achieves the best performance compared to existing methods, suggesting that the knowledge of subway and ride-hailing demand can indeed benefit the demand prediction of bike sharing. Further analysis demonstrates the effectiveness of all the proposed components and the interpretability of our proposed model.

In future works, this research can be improved in the following aspects. Firstly, in this research, we only experiment with station-based shared bikes due to lack of data. Future research can apply our proposed model to stationless BSS and other transport modes. Secondly, while our proposed MR-GNNExplainer can provide local interpretation for the GNN's prediction of each node, it would be meaningful to provide global interpretation of the entire multi-relational graph so that the characteristics of multimodal networks can be better understood. This might be achieved by extending more advanced explainable GNN techniques such as PGExplainer \cite{luo2020parameterized}. Thirdly, future research can leverage cross-mode information for long-term bike sharing demand prediction. For example, in cases of service expansion, there is no existing bike sharing demand data in the target region. In such cases, we can potentially utilize existing demand data of other modes to estimate the likely usage of bike sharing.


\bibliographystyle{IEEEtran}
\bibliography{ref}

\begin{thebibliography}{10}
\providecommand{\url}[1]{#1}
\csname url@samestyle\endcsname
\providecommand{\newblock}{\relax}
\providecommand{\bibinfo}[2]{#2}
\providecommand{\BIBentrySTDinterwordspacing}{\spaceskip=0pt\relax}
\providecommand{\BIBentryALTinterwordstretchfactor}{4}
\providecommand{\BIBentryALTinterwordspacing}{\spaceskip=\fontdimen2\font plus
\BIBentryALTinterwordstretchfactor\fontdimen3\font minus
  \fontdimen4\font\relax}
\providecommand{\BIBforeignlanguage}[2]{{%
\expandafter\ifx\csname l@#1\endcsname\relax
\typeout{** WARNING: IEEEtran.bst: No hyphenation pattern has been}%
\typeout{** loaded for the language `#1'. Using the pattern for}%
\typeout{** the default language instead.}%
\else
\language=\csname l@#1\endcsname
\fi
#2}}
\providecommand{\BIBdecl}{\relax}
\BIBdecl

\bibitem{xu2018station}
C.~Xu, J.~Ji, and P.~Liu, ``The station-free sharing bike demand forecasting
  with a deep learning approach and large-scale datasets,''
  \emph{Transportation research part C: emerging technologies}, vol.~95, pp.
  47--60, 2018.

\bibitem{jin2020dockless}
K.~Jin, W.~Wang, S.~Li, P.~Liu, and H.~Sun, ``Dockless shared-bike demand
  prediction with temporal convolutional networks,'' in \emph{CICTP 2020},
  2020, pp. 2851--2863.

\bibitem{geng2019spatiotemporal}
X.~Geng, Y.~Li, L.~Wang, L.~Zhang, Q.~Yang, J.~Ye, and Y.~Liu, ``Spatiotemporal
  multi-graph convolution network for ride-hailing demand forecasting,'' in
  \emph{Proceedings of the AAAI conference on artificial intelligence},
  vol.~33, no.~01, 2019, pp. 3656--3663.

\bibitem{yu2017spatio}
B.~Yu, H.~Yin, and Z.~Zhu, ``Spatio-temporal graph convolutional networks: A
  deep learning framework for traffic forecasting,'' \emph{arXiv preprint
  arXiv:1709.04875}, 2017.

\bibitem{zhang2018short}
C.~Zhang, L.~Zhang, Y.~Liu, and X.~Yang, ``Short-term prediction of
  bike-sharing usage considering public transport: A lstm approach,'' in
  \emph{2018 21st International Conference on Intelligent Transportation
  Systems (ITSC)}.\hskip 1em plus 0.5em minus 0.4em\relax IEEE, 2018, pp.
  1564--1571.

\bibitem{ye2019co}
J.~Ye, L.~Sun, B.~Du, Y.~Fu, X.~Tong, and H.~Xiong, ``Co-prediction of multiple
  transportation demands based on deep spatio-temporal neural network,'' in
  \emph{Proceedings of the 25th ACM SIGKDD International Conference on
  Knowledge Discovery \& Data Mining}, 2019, pp. 305--313.

\bibitem{wang2020multi}
S.~Wang, H.~Miao, H.~Chen, and Z.~Huang, ``Multi-task adversarial
  spatial-temporal networks for crowd flow prediction,'' in \emph{Proceedings
  of the 29th ACM international conference on information \& knowledge
  management}, 2020, pp. 1555--1564.

\bibitem{xu2022adaptive}
H.~Xu, T.~Zou, M.~Liu, Y.~Qiao, J.~Wang, and X.~Li, ``Adaptive spatiotemporal
  dependence learning for multi-mode transportation demand prediction,''
  \emph{IEEE Transactions on Intelligent Transportation Systems}, 2022.

\bibitem{cho2021enhancing}
J.-H. Cho, S.~W. Ham, and D.-K. Kim, ``Enhancing the accuracy of peak hourly
  demand in bike-sharing systems using a graph convolutional network with
  public transit usage data,'' \emph{Transportation Research Record}, vol.
  2675, no.~10, pp. 554--565, 2021.

\bibitem{li2021multi}
C.~Li, L.~Bai, W.~Liu, L.~Yao, and S.~T. Waller, ``A multi-task memory network
  with knowledge adaptation for multimodal demand forecasting,''
  \emph{Transportation Research Part C: Emerging Technologies}, vol. 131, p.
  103352, 2021.

\bibitem{lv2021mobility}
Y.~Lv, D.~Zhi, H.~Sun, and G.~Qi, ``Mobility pattern recognition based
  prediction for the subway station related bike-sharing trips,''
  \emph{Transportation Research Part C: Emerging Technologies}, vol. 133, p.
  103404, 2021.

\bibitem{hua2022transfer}
M.~Hua, F.~C. Pereira, Y.~Jiang, and X.~Chen, ``Transfer learning for
  cross-modal demand prediction of bike-share and public transit,'' \emph{arXiv
  preprint arXiv:2203.09279}, 2022.

\bibitem{liang2021joint}
Y.~Liang, G.~Huang, and Z.~Zhao, ``Joint demand prediction for multimodal
  systems: A multi-task multi-relational spatiotemporal graph neural network
  approach,'' \emph{Transportation Research Part C: Emerging Technologies},
  vol. 140, p. 103731, 2022.

\bibitem{simini2021deep}
F.~Simini, G.~Barlacchi, M.~Luca, and L.~Pappalardo, ``A deep gravity model for
  mobility flows generation,'' \emph{Nature communications}, vol.~12, no.~1,
  pp. 1--13, 2021.

\bibitem{zhao2022deep}
Z.~Zhao and Y.~Liang, ``Deep inverse reinforcement learning for route choice
  modeling,'' \emph{arXiv preprint arXiv:2206.10598}, 2022.

\bibitem{ying2019gnnexplainer}
Z.~Ying, D.~Bourgeois, J.~You, M.~Zitnik, and J.~Leskovec, ``Gnnexplainer:
  Generating explanations for graph neural networks,'' \emph{Advances in neural
  information processing systems}, vol.~32, 2019.

\bibitem{liang2022bike}
Y.~Liang, G.~Huang, and Z.~Zhao, ``Bike sharing demand prediction based on
  knowledge sharing across modes: A graph-based deep learning approach,''
  \emph{accepted by 2022 IEEE 25th international conference on intelligent
  transportation systems}, 2022.

\bibitem{yoon2012cityride}
J.~W. Yoon, F.~Pinelli, and F.~Calabrese, ``Cityride: a predictive bike sharing
  journey advisor,'' in \emph{2012 IEEE 13th International Conference on Mobile
  Data Management}.\hskip 1em plus 0.5em minus 0.4em\relax IEEE, 2012, pp.
  306--311.

\bibitem{rudloff2014modeling}
C.~Rudloff and B.~Lackner, ``Modeling demand for bikesharing systems:
  neighboring stations as source for demand and reason for structural breaks,''
  \emph{Transportation Research Record}, vol. 2430, no.~1, pp. 1--11, 2014.

\bibitem{feng2018hierarchical}
S.~Feng, H.~Chen, C.~Du, J.~Li, and N.~Jing, ``A hierarchical demand prediction
  method with station clustering for bike sharing system,'' in \emph{2018 IEEE
  Third International Conference on Data Science in Cyberspace (DSC)}.\hskip
  1em plus 0.5em minus 0.4em\relax IEEE, 2018, pp. 829--836.

\bibitem{guidon2020expanding}
S.~Guidon, D.~J. Reck, and K.~Axhausen, ``Expanding a (n)(electric)
  bicycle-sharing system to a new city: Prediction of demand with spatial
  regression and random forests,'' \emph{Journal of Transport Geography},
  vol.~84, p. 102692, 2020.

\bibitem{zhou2018predicting}
X.~Zhou, Y.~Shen, Y.~Zhu, and L.~Huang, ``Predicting multi-step citywide
  passenger demands using attention-based neural networks,'' in
  \emph{Proceedings of the Eleventh ACM International Conference on Web Search
  and Data Mining}, 2018, pp. 736--744.

\bibitem{lin2018predicting}
L.~Lin, Z.~He, and S.~Peeta, ``Predicting station-level hourly demand in a
  large-scale bike-sharing network: A graph convolutional neural network
  approach,'' \emph{Transportation Research Part C: Emerging Technologies},
  vol.~97, pp. 258--276, 2018.

\bibitem{wu2019graph}
Z.~Wu, S.~Pan, G.~Long, J.~Jiang, and C.~Zhang, ``Graph wavenet for deep
  spatial-temporal graph modeling,'' \emph{arXiv preprint arXiv:1906.00121},
  2019.

\bibitem{wang2022nonlinear}
Y.~Wang, Z.~Zhan, Y.~Mi, A.~Sobhani, and H.~Zhou, ``Nonlinear effects of
  factors on dockless bike-sharing usage considering grid-based spatiotemporal
  heterogeneity,'' \emph{Transportation Research Part D: Transport and
  Environment}, vol. 104, p. 103194, 2022.

\bibitem{farahani2021brief}
A.~Farahani, S.~Voghoei, K.~Rasheed, and H.~R. Arabnia, ``A brief review of
  domain adaptation,'' \emph{Advances in data science and information
  engineering}, pp. 877--894, 2021.

\bibitem{liu2021knowledge}
Y.~Liu, B.~Guo, D.~Zhang, D.~Zeghlache, J.~Chen, K.~Hu, S.~Zhang, D.~Zhou, and
  Z.~Yu, ``Knowledge transfer with weighted adversarial network for cold-start
  store site recommendation,'' \emph{ACM Transactions on Knowledge Discovery
  from Data (TKDD)}, vol.~15, no.~3, pp. 1--27, 2021.

\bibitem{tzeng2014deep}
E.~Tzeng, J.~Hoffman, N.~Zhang, K.~Saenko, and T.~Darrell, ``Deep domain
  confusion: Maximizing for domain invariance,'' \emph{arXiv preprint
  arXiv:1412.3474}, 2014.

\bibitem{long2015learning}
M.~Long, Y.~Cao, J.~Wang, and M.~Jordan, ``Learning transferable features with
  deep adaptation networks,'' in \emph{International conference on machine
  learning}.\hskip 1em plus 0.5em minus 0.4em\relax PMLR, 2015, pp. 97--105.

\bibitem{ganin2015unsupervised}
Y.~Ganin and V.~Lempitsky, ``Unsupervised domain adaptation by
  backpropagation,'' in \emph{International conference on machine
  learning}.\hskip 1em plus 0.5em minus 0.4em\relax PMLR, 2015, pp. 1180--1189.

\bibitem{tang2022domain}
Y.~Tang, A.~Qu, A.~H. Chow, W.~H. Lam, S.~Wong, and W.~Ma, ``Domain adversarial
  spatial-temporal network: A transferable framework for short-term traffic
  forecasting across cities,'' \emph{arXiv preprint arXiv:2202.03630}, 2022.

\bibitem{fangtransfer}
Z.~Fang, D.~Wu, L.~C. Lu~Pan, and Y.~Gao, ``When transfer learning meets
  cross-city urban flow prediction: Spatio-temporal adaptation matters.''

\bibitem{yuan2020xgnn}
H.~Yuan, J.~Tang, X.~Hu, and S.~Ji, ``Xgnn: Towards model-level explanations of
  graph neural networks,'' in \emph{Proceedings of the 26th ACM SIGKDD
  International Conference on Knowledge Discovery \& Data Mining}, 2020, pp.
  430--438.

\bibitem{luo2020parameterized}
D.~Luo, W.~Cheng, D.~Xu, W.~Yu, B.~Zong, H.~Chen, and X.~Zhang, ``Parameterized
  explainer for graph neural network,'' \emph{Advances in neural information
  processing systems}, vol.~33, pp. 19\,620--19\,631, 2020.

\bibitem{li2022survey}
Y.~Li, J.~Zhou, S.~Verma, and F.~Chen, ``A survey of explainable graph neural
  networks: Taxonomy and evaluation metrics,'' \emph{arXiv preprint
  arXiv:2207.12599}, 2022.

\bibitem{zhou2021modeling}
Q.~Zhou, J.~Gu, X.~Lu, F.~Zhuang, Y.~Zhao, Q.~Wang, and X.~Zhang, ``Modeling
  heterogeneous relations across multiple modes for potential crowd flow
  prediction,'' in \emph{Proceedings of the AAAI Conference on Artificial
  Intelligence}, vol.~35, no.~5, 2021, pp. 4723--4731.

\bibitem{liebel2018auxiliary}
L.~Liebel and M.~K{\"o}rner, ``Auxiliary tasks in multi-task learning,''
  \emph{arXiv preprint arXiv:1805.06334}, 2018.

\end{thebibliography}

\vskip 0pt plus -1fil
\begin{IEEEbiography}[{\includegraphics[width=1in,height=1.25in,clip,keepaspectratio]{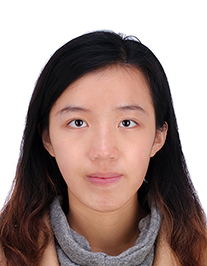}}]{Yuebing Liang}
is currently pursuing the Ph.D. degree at the Department of Urban Planning and Design, The University of Hong Kong, Hong Kong. She received her Master's and Bachelor's degrees from Tsinghua University. Her research interests include urban data science and intelligent transport systems.
\end{IEEEbiography}
\vskip 0pt plus -1fil
\begin{IEEEbiography}[{\includegraphics[width=1in,height=1.25in,clip,keepaspectratio]{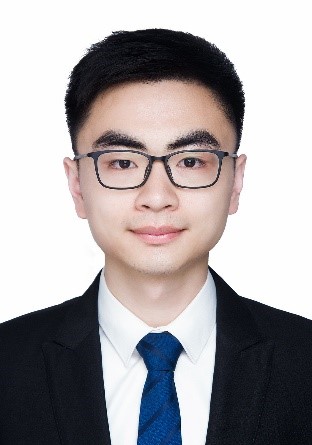}}]{Guan Huang}
is currently pursuing the Ph.D. degree at the Department of Urban Planning and Design, The University of Hong Kong, Hong Kong. He received his Master's and Bachelor's degrees from Wuhan University. His research interests include urban data science and shared mobility.
\end{IEEEbiography}
\vskip 0pt plus -1fil
\begin{IEEEbiography}[{\includegraphics[width=1in,height=1.25in,clip,keepaspectratio]{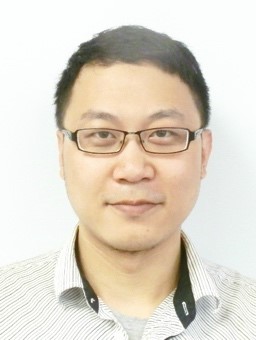}}]{Zhan Zhao} is an Assistant Professor in Department of Urban Planning and Design, The University of Hong Kong (HKU), and also affiliated with HKU Musketeers Foundation Institute of Data Science. He holds a Ph.D. degree from the Massachusetts Institute of Technology, a Master's degree from the University of British Columbia, and a Bachelor's degree from Tongji University. His research interests include human mobility modeling, public transportation systems, and urban data science.
\end{IEEEbiography}





\end{document}